\documentclass[%
 preprint,
 amsmath,amssymb
 prb,
  superscriptaddress,
unsortedaddress,
]{revtex4-2}

\usepackage{mhchem}
\usepackage[utf8]{inputenc} 
\usepackage[T1]{fontenc}    
\usepackage{hyperref}       
\usepackage{url}            
\usepackage{booktabs}       
\usepackage{amsfonts}       
\usepackage{nicefrac}       
\usepackage{microtype}      
\usepackage{makecell}
\usepackage{amsmath}
\usepackage{amssymb}
\usepackage{color}
\usepackage{epstopdf}
\usepackage{physics}
\usepackage{enumitem}
\usepackage{graphicx}
\usepackage{subcaption}
\usepackage{dcolumn}
\usepackage{bm}
\usepackage[yyyymmdd, 24hr]{datetime}

\begin{document}
\author{Tomohiko Konno}
\email[Corresponding author: ]{tomohiko@nict.go.jp}
\affiliation{National Institute of Information and Communications Technology}
\title{Deep-Learning Estimation of Band Gap with the Reading-Periodic-Table Method and Periodic Convolution Layer} 
\begin{abstract}
We verified that the deep learning method named reading periodic table introduced by Deep Learning Model for Finding New Superconductors (Ref.~\cite{konno2018deep}), which utilizes deep learning to read the periodic table and the laws of the elements, is applicable not only for superconductors, for which the method was originally applied, but also for other problems of materials by demonstrating band gap estimations. We then extended the method to learn the laws better by directly learning the cylindrical periodicity between the right- and left-most columns in the periodic table at the learning representation level, that is, by considering the left- and right-most columns to be adjacent to each other
. Thus, while the original method handles the table as is, the extended method treats the periodic table as if its two edges are connected. This is achieved using novel layers named periodic convolution layers, which can handle inputs exhibiting periodicity and may be applied to other problems related to computer vision, time series, and so on for data that possess some periodicity. In the reading periodic table method, no material feature or descriptor is required as input. We demonstrated two types of deep learning estimation: methods to estimate the existence of a band gap, and methods to estimate the value of the band gap given when the existence of the band gap in the materials is known. Finally, we discuss the limitations of the dataset and model evaluation method. We may be unable to distinguish good models based on the random train--test split scheme; thus, we must prepare an appropriate dataset where the training and test data are temporally separate. The code and the data are open.
\end{abstract}
\maketitle
\newpage
\section{Introduction.} Machine learning (ML) methods have been employed in the search for inorganic materials~\cite{konno2018deep,doi:10.1021/acs.jpclett.8b00009}, and useful libraries have been developed~\cite{ong2013python,ward2016general}. In the case of organic materials, machine learning methods have been studied, typically by employing graph structures~\cite{Zhang2018DeepLO,Zhou2018GraphNN}, which are relevant to problems in computer science. Researchers have invested more efforts into searching for organic materials than for inorganic ones. However, the search for inorganic materials still has a wide scope. Density functional theory (DFT)~\cite{doi:10.1021/cr200107z,RevModPhys.87.897,doi:10.1080/00268976.2017.1333644}, which involves first-principle computational calculations, requires expensive computational resources, is difficult to apply to highly correlated systems, and often requires ordered crystal structures, although progress is being made to address such issues. Machine learning and DFT can be used to complement each other, and both must be investigated further. Deep learning~\cite{Murphy:2012:MLP:2380985,goodfellow2016deep,zhang2019dive} has achieved advances in image recognition~\cite{krizhevsky2012imagenet}, machine translation~\cite{Vaswani2017AttentionIA}, image generation~\cite{goodfellow2014generative}, natural language inference~\cite{peters2018deep,devlin2018bert}, raw audio generation \cite{oord2016wavenet}, and imperfect information games~\cite{Blair864}. Now, deep learning is finding increasing applications in mathematics and physics as well, in the fields of quantum systems~\cite{doi:10.7566/JPSJ.89.022001}, particle physics~\cite{Bourilkov2019MachineAD}, Gauss-Manin connection~\cite{Heal2020DeepLG}, neural networks and quantum field theory~\cite{Halverson_2021}, holographic QCD~\cite{Hashimoto2020NeuralOA}, conformal field theory~\cite{Chen2020MachineLE}, black hole metrics~\cite{Yan2020DeepLB}, supergravity~\cite{Krishnan2020MachineL}, Seiberg-Witten curves~\cite{He2020MachineLearningDD}, Calabi-Yau manifolds~\cite{he2020machine,Erbin2020MachineLF}, etc. Deep learning possesses superior capabilities over prior machine learning methods, such as support vector machines (SVM) \cite{Boser:1992:TAO:130385.130401,Cortes1995}, random forest \cite{Breiman2001}, and k-means \cite{zbMATH03129892,macqueen1967,10009668881}.

A deep learning method for finding new superconductors was introduced in our previous study~\cite{konno2018deep} (Although it is not technically deep learning, the critical temperature of superconductors was studied by the random forest method, which is a method of machine learning, in ref~\cite{stanev2018machine}). In the study~\cite{konno2018deep}, a deep learning model was trained to read the periodic table to learn the law of elements in order to identify novel materials not present in the data. This technique does not require calculations for inputting any feature or material descriptor. The method was named \emph{reading periodic table}. This deep learning model outperformed random forest.

Nevertheless, two problems remain. The deep learning method of \emph{reading periodic table} does not use any specific feature of superconductors; it only uses the periodic table. Although it was demonstrated that the method was useful for problems of superconductors, it is still unclear whether the method can be used to solve other material-related problems. If it can be used, this will open significant opportunities. Thus, we verify it in the present paper. The second problem is regarding the cylindrical periodicity between the left— and right—most columns in the table. The periodic table is effectively cylindrical, in contrast to its two-dimensional appearance; the left-most and right-most columns are adjacent. Our previous method was not designed to allow the deep learning model to read the periodic table in the cylindrical form at the level of training representation. If the layers are sufficiently deep, neural networks can also learn the laws from the cylindrical periodicity without making any specific modifications to the original method of \emph{reading periodic table}. However, they may learn them insufficiently well, and it is unclear whether the deep learning model learns the laws represented by the cylindrical periodicity, as deep learning is a black box model. Hence, it is better to use a learning representation that allows the deep learning model to unquestionably learn the laws represented by the cylindrical periodicity. We extended our previous method \cite{konno2018deep} to reflect the cylindrical periodicity at the learning representation level. We designed this functionality as a new layer named the \emph{periodic convolution layer}, which performs the required operation, i.e., the layers process the periodic table as if the table is cylindrically rolled-up, even though the input is the ordinary two dimensional periodic table. This method is named reading periodic table with cylindrical periodicity. The \emph{periodic convolution layer} can be used for other problems related to computer vision, time series data, and so on, if the data being examined contains some periodicity.

To solve the two aforementioned problems, we demonstrate that the extended method can predict band gaps. A band gap is a fundamental material property that forms the basis for separation among conductors, semiconductors, and insulators; it influences thermal and electrical conductivities, the functioning of light emitting and laser diodes, etc. In addition, its structure is relevant to some topological matter that has recently been reported and is receiving considerable attention. To design functional materials, knowledge of the band gap is important. ML-based band-gap estimation has already been performed in earlier studies \cite{gu2006using,zhaochun1998artificial,dey2014informatics,lee2016prediction,pilania2016machine,Zhuo2018PredictingTB}. We demonstrate that the proposed method of \emph{reading periodic table} with cylindrical periodicity has estimation capabilities comparable to those of SVM classification and regression studied in Ref.~\cite{Zhuo2018PredictingTB}. We also highlight the problem in the random train-test-split model evaluation scheme, in that we may not be able to identify good models by using the scheme. 

The main contributions of the paper are as follows. In the previous study, we introduced the method named reading periodic table, which uses deep learning learn to read the periodic table and estimate the critical temperature of superconductors. In this paper, we extended the method for deep learning to also learn the cylindrical periodicity directly, i.e. by considering the fact that the right- and left-most columns are adjacent in the table. We then demonstrated that the method has wide applicability to material-related problems other than superconductors by demonstrating two band gap estimations. This verification is necessary for future applications.

\begin{figure}[htbp]
        \centering
        \includegraphics[width=\textwidth]{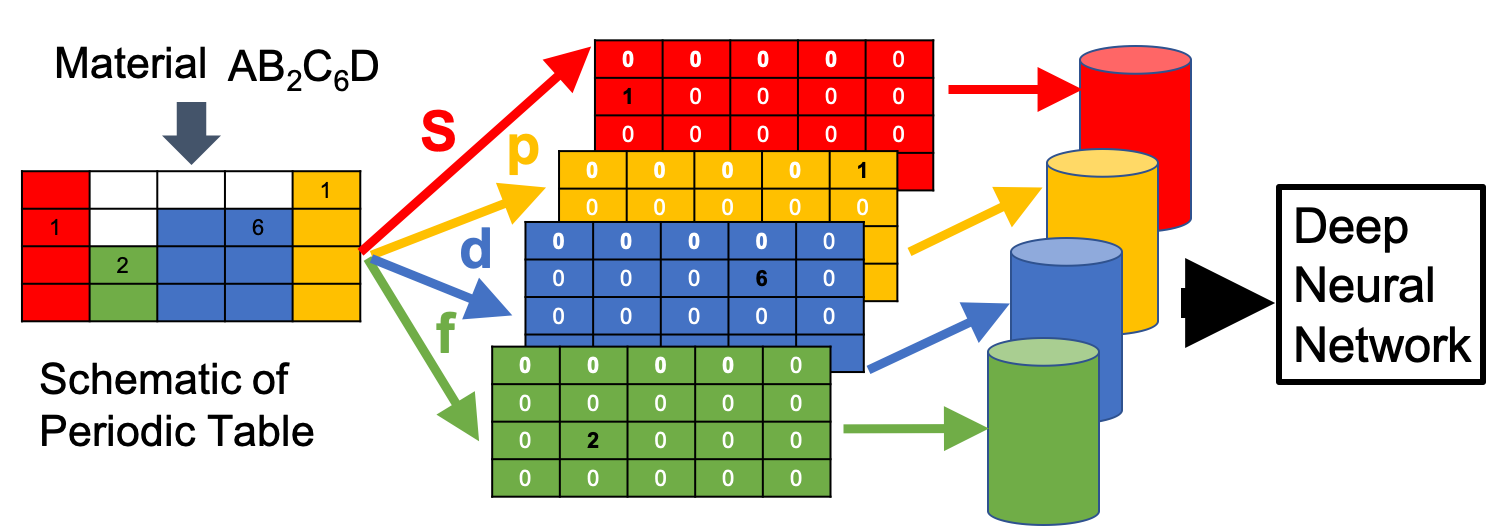}
            \caption{Reading periodic table with cylindrical periodicity (color online). \\
            The composite values of a hypothetical material \ce{AB2C6D} are listed in the periodic table, which is divided into four separate tables according to the electron orbitals. In the middle four tables, as an example, for the second top table, all the values on the table except for those of the p-block elements are $0$.
Then, the tables are processed as if they are rolled up in the horizontal direction. This is done to allow the deep learning model to learn that the left- and right-most columns of the table are adjacent at the learning representation level.}\label{fig: periodicity_behind}
\end{figure}

\begin{figure}[htbp]
    \begin{subfigure}[t]{0.7\textwidth}
        \centering
        \includegraphics[width=1\textwidth]{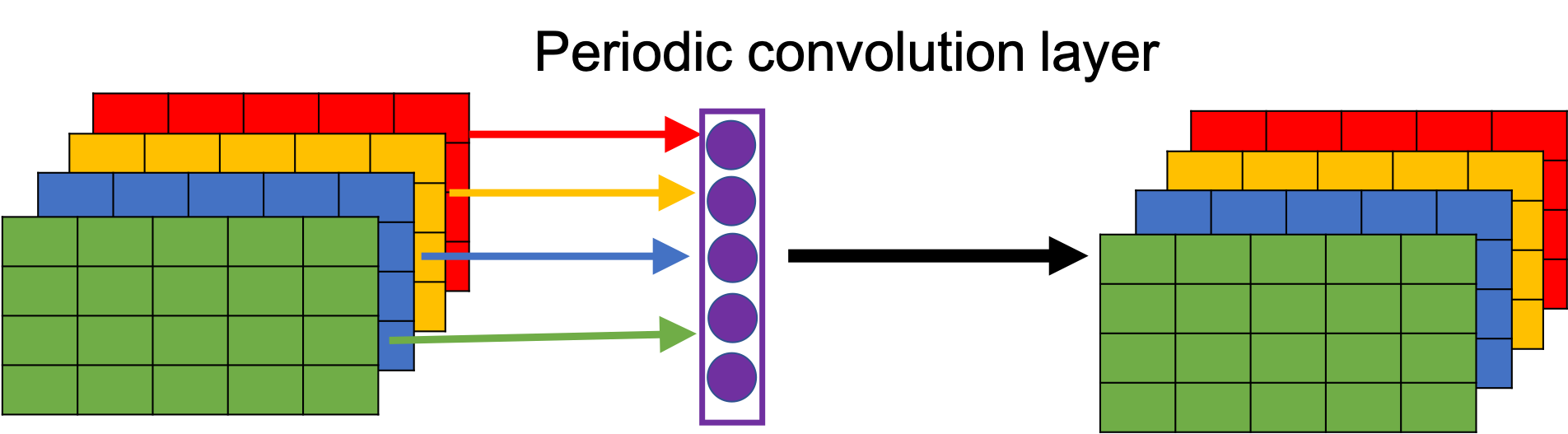}
    \end{subfigure} 
    \begin{subfigure}[t]{0.7\textwidth}
        \centering
        \includegraphics[width=1\textwidth]{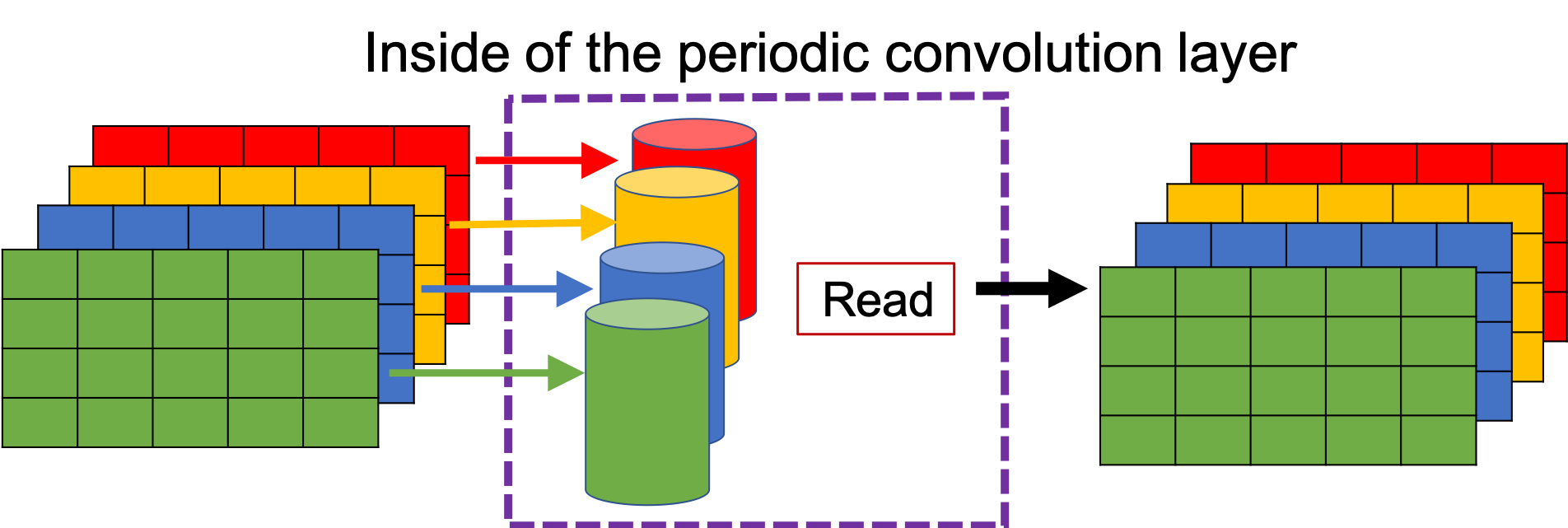}
	\end{subfigure}
	\caption{Periodic convolution layer.
Top: schematic of the periodic convolution layer that reads the periodic table with cylindrical periodicity (color online). Bottom: inside of the periodic convolution layer. If the learning representation of the 2D periodic table is input to the layer, the layer considers the 2D periodic table with four orbitals as if the left- and right-most columns are adjacent and the tables are cylindrically rolled up. The output of the layer is in 2D $\times$ arbitrary depth form, which can be addressed in the subsequent layers; its input depths can also be arbitrary.}\label{fig: behind_periodicity_layer}
\end{figure}

\section{Method} The learning representation of the \emph{reading periodic table} allows the deep learning model to read the periodic table and learn the laws of elements to identify novel materials. The method is designed to learn the relative positions of elements on periodic table, which is difficult when using a 118-dim one-hot representation. Our methods allow us to extract the laws represented by the periodic table. A material feature or descriptor is not required, and instead, the features are automatically determined from the periodic table via deep learning. Furthermore, it is not necessary to input the crystal structure of the materials. This is both advantageous and disadvantageous. The advantage is as follows.
Because we do not need to input spatial information, we just need to  input the chemical formula only to estimate material properties We do not have to calculate the spatial structure of new materials via first principle calculations, which require high computational cost, before inputting spatial structure to the machine learning model. Furthermore, we could not acquire spatial information for the experimental data on band gaps. 
Databases of experimentally measured spatial information are available, such as the Crystallography Open Database (COD)~\cite{gravzulis2011crystallography,gravzulis2009crystallography,Downs2003}. \href{http://www.crystallography.net/cod/}{COD} is an open-access collection of the crystal structures of organic, inorganic, and metal-organic materials, and it has approximately 460,000 entries as of 2020. However, there are no databases for specific problems like band gaps, superconductors, and so forth. It takes significant effort and time to determine the spatial structure of the experimental data of materials related to band gap problems using such a database. In such databases, the same chemical formulas can have different spatial structures. To determine the spatial structure of a specific material, we need to check original papers. The disadvantage of not inputting the crystal structure is that the capability of machine learning will improve if we use the complete information of spatial structures, since it plays an important role in physics.
 Although it is not visibly apparent, the left- and right-most columns of the periodic table are adjacent, and it also possesses cylindrical periodicity. Thus, for the above-mentioned reasons, we extend the method so that deep learning can also learn the laws represented by the cylindrical periodicity more directly at the learning representation level, as illustrated in Fig.~\ref{fig: periodicity_behind}. We achieve this functionality using a new layer named \emph{periodic convolution layer}, as illustrated in Fig~\ref{fig: behind_periodicity_layer}. The layer can be applied to other problems that use input data possessing some periodicity.
\subsection{Hyperparameters}
For both estimation tasks, we used the absolute representation in the method of reading periodic table with cylindrical periodicity, where the learning rate is $10^{-4}$, Adam is used as the optimizer, and a network with a depth of 64 is used. In deep learning, a depth of 64 is not so deep compared to that in modern neural network architectures.

In the absolute representation, \ce{H2O} is represented as {H: 2, O:1}, whereas in its relative representation, it is {H:$\frac{2}{3}$, O: $\frac{1}{3}$}. Except for the fact that that the representation is the absolute one, the hyperparameters are identical to those used in our previous study~\cite{konno2018deep}; the learning rate is $10^{-4}$, and the epochs are 200 for both band gap existence estimation and gap value estimation. Parameter tuning may improve the scores.

\subsection{Neural Network Structure}
\textit{Basic periodic block.} To explain the model structure, we need to describe the basic periodic block. Let a basic periodic block be denoted by basic periodic block[input channel, inside channel, output channel, kernel size], which is composed of the following layers. First, we have the periodic convolution[input channel, output channel=inside channel, kernel size], followed by batch norm, followed by a ReLU, followed by periodic convolution[input channel=inside channel, output channel, kernel size], followed by batch norm, and then followed by a ReLU. We also have a skip connection in the basic periodic block between input of the block and before the last ReLU.

The structure of the neural network is as follows. First, we have the periodic convolution[input channel=4, output channel=10, kernel size=(1,1)]  Then, ten basic periodic blocks[input channel=10, inside channel=10, output channel=10, kernel size=(3,3)] follow. Next, we have conv[input channel=10, output channel=20, kernel size=(2, 4), stride=(1, 2), padding=(0, 3)] followed by batch norm and a ReLU. Then, basic periodic block[input channel=20, inside channel=20, output channel=20, kernel size=(3,3)] follows. Next, we have conv[input channel=20, output channel=30, kernel size=(2, 4), stride=(1, 2), padding=(0, 3)] followed by batch norm and a ReLU. Then, basic periodic block[input channel=30, inside channel=30, output channel=30, kernel size=(3,3)] follows. Next, we have conv[input channel=30, output channel=40, kernel size=(2, 4), stride=(1, 2), padding=(0, 3)] followed by batch norm and a ReLU. Then, basic periodic block[input channel=40, inside channel=40, output channel=40, kernel size=(3,3)] follows. Next, we have conv[input channel=40, output channel=50, kernel size=(2, 4), stride=(1, 2), padding=(0, 3)] followed by batch norm and a ReLU. Then, basic periodic block[input channel=50, inside channel=50, output channel=50, kernel size=(3,3)] follows. Next, we have conv[input channel=50, output channel=60, kernel size=(2, 4), stride=(1, 2), padding=(0, 3)] followed by batch norm and a ReLU. Then, basic periodic block[input channel=60, inside channel=60, output channel=60, kernel size=(3,3)] follows. Next, we have conv[input channel=60, output channel=70, kernel size=(2, 4), stride=(1, 2), padding=(0, 3)] followed by batch norm and a ReLU. Then, basic periodic block[input channel=70, inside channel=70, output channel=70, kernel size=(3,3)] follows. Finally, a full connection layer FC[input=210, output=100], FC[input=100, output=100], and [input=100, output=1] follows.

\section{Demonstration of the applicability of the method}
\subsection{Binary Classification and Regression of Band Gaps.}
Here, we demonstrate two types of estimations: (1) estimating the existence of a band gap and predicting whether the material is a nonmetal, and (2) estimating the values of the band gap given that the existence of a band gap in the materials is known in advance. We basically used the same dataset employed in the previous study using SVM~\cite{Zhuo2018PredictingTB} with correction to compare our results with the SVM-based results.
It is noteworthy that the previous study used an existing method, SVM, whereas we invented a novel method.

\subsection{Summary of the Data}
We use the experimental data of 3734 gapped materials, of which 2473 are unique compositions. For the binary classification, we also use the first principle calculation data of 2450 non-gapped materials from Material Project~\cite{Jain2013}. The numbers are balanced to avoid possible bias caused by an imbalance between the number of gapped and non-gapped materials. Only unique compositions are used. We use only the experimental data for the regression of band gaps. We use all the experimental data of the 3734 gapped materials to accord with previous studies.
\subsection{Classification of band gap existence.}
\begin{figure}[htbp]
    \begin{center}
        \includegraphics[width=.6\hsize]{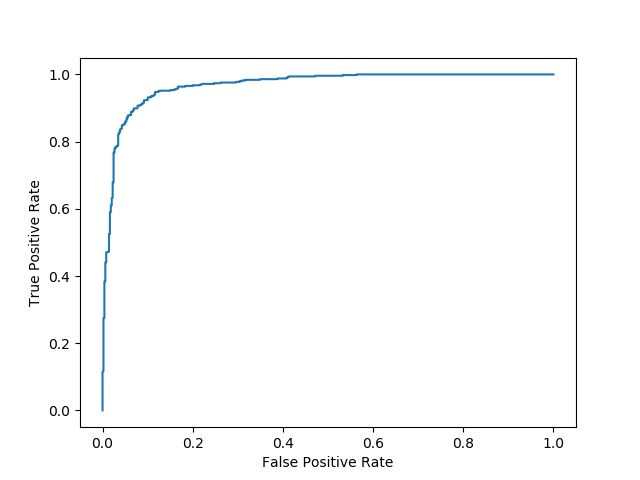}
        \end{center}
            \caption{ROC curve for the binary classification of band-gap existence for test data (color online)}\label{fig: roc}
    \end{figure}

We randomly split the dataset into training and test data in a ratio of 80 to 20. The results of the binary classification are summarized in Table~\ref{tab: binary}, and the ROC curve is illustrated in Fig.~\ref{fig: roc}.

\begin{table}
\centering
\begin{tabular}{|c|c|c|c|c|}
\hline
accuracy & precision & recall & f1 & AUC \\ \hline
0.92 & 0.91 & 0.94 & 0.93 & 0.97 \\ \hline
\end{tabular}
\caption{Binary classification of band gap for test data}\label{tab: binary}
\end{table}
Our model exhibits good scores that are comparable to those obtained in the previous study, where the accuracy = 0.92, precision = 0.89, recall = 0.95, f1 = 0.92, and area under curve (AUC) = 0.97. We will compare our result to that of previous study. The AUCs are the same. In four basic metrics (accuracy, precision, recall, and f1), the previous model is only better in terms of recall, and the accuracies are the same. Our model is better in terms of precision and f1. f1 is particularly important because it is an integrated measure of precision and recall.

\subsection{Regression of Band-Gap Values.} 
\begin{figure}[htbp]
    \begin{center}
        \includegraphics[width=.7\hsize]{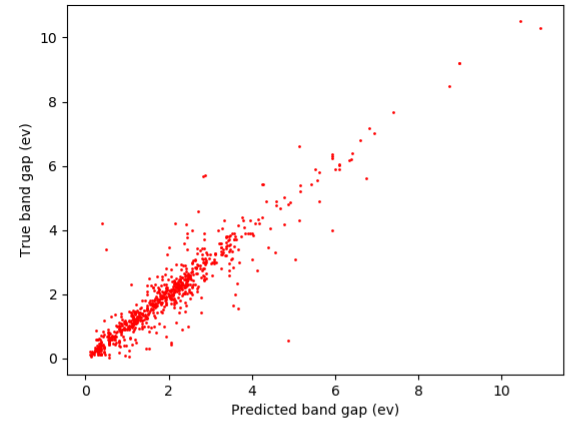}
        \caption{Scatter plot for the band gap regression: predicted vs. true band gap values for test data (color online)}\label{fig: scatter}
        \end{center}
    \end{figure}

The dataset used had 3734 compositions. We did not remove identical compositions with different band gaps to accord with the previous study, as they may differ in their crystal structures, which could not be understood from the available data. We also randomly separated the dataset into training and test data in a ratio of 80 to 20. The results of band-gap regression (for test data) are summarized in Table~\ref{tab: reg}, and the scatter plot is illustrated in Fig.~\ref{fig: scatter}. $R^{2} =$ 0.92, and the root-mean squared error (RMSE) = 0.42. Our model achieved good scores that are comparable to those obtained in the previous study, where $R^{2}=$ 0.90 and RMSE = 0.45. Our model outperforms the previous model in terms of both R2 and RMSE scores. Two machine learning methods predict different values for one instance (material), by which we mean a candidate material. Different models can be useful for prediction. More discussion on regression and classification is given in supplementary material~\cite{supple}. 

\begin{table}
\centering
\begin{tabular}{|c|c|}
\hline
R-squared & RMSE\\ \hline
0.92 & 0.42 \\ \hline
\end{tabular}
\caption{Regression of band gap for test data. }\label{tab: reg}
\end{table}


%

\section{Conclusion}

In this study, we extended the deep learning--based reading periodic table method, which reads the periodic table and learns the laws of elements. We solved two problems. First, we determined whether we can use the deep learning method of reading periodic table~\cite{konno2018deep} not only for superconductors but also for other material-related problems. Second, we used the cylindrical periodicity between the left- and right-most columns of the periodic table directly.
 
Through the extensions made to the deep learning--based reading periodic table method, it can now also learn the laws represented by the cylindrical periodicity of the periodic table at the learning representation level, meaning that the left- and right-most columns of the table are adjacent. We implemented this functionality through a new layer named \emph{periodic convolution layer}. Even if the input to the layer is in the form of an ordinary 2D periodic table, the layer considers the table as if it were rolled up horizontally, and the output is also in a 2D form with arbitrary depth so that it is handled by the succeeding layers in a similar manner. The periodic convolution layer can also be applied to other problems related to computer vision, time series data, and so on, provided that the input data have some periodicity.

We demonstrated the applicability of the method for other material-related problems by solving two problems related to the band gap. The deep learning method satisfactorily provides a binary prediction of band-gap existence and a regression of the band-gap values. The training and test data obtained by randomly splitting the dataset were used to compare our results with those of the previous studies. However, because the test data is inevitably similar to the training data, the estimation by machine learning with the test data may become very easy. To more accurately evaluate models, we require an appropriately prepared dataset. Thus, the data must be divided temporally, such that the data until a certain year constitute the training data and the data after that year constitute the test data, as discussed in Ref.\cite{konno2018deep}. This is the same situation under which we will use machine learning models for material search, and the evaluation of the machine learning models must be performed under similar conditions. However, as we could not know when the data was obtained, we could not temporally separate the dataset in this study. Contrary to the naive intuition of a machine learning novice, the preparation of an appropriate dataset is very difficult and is among the most crucial contributions to the progress of the field. We would like to stress that the development of such a dataset remains a challenge for the machine learning community, and if the dataset is appropriately prepared, we would be more capable of distinguishing and making good models based on such a dataset.

\section*{The code and the data}
The data used for band gap existence binary classification, the data used for the band gap regression, the code for the periodic convolution layer, the code for the neural network will be found in the \href{https://github.com/tomo835g/Superconductors}{link}, since a link cannot be used in a paper in this journal. Readers can use the code and the data provided that they cite this paper. Furthermore, the code that transforms chemical formulas into the reading periodic table data format can be obtained under the same condition. See the conditions in detail from the \href{https://github.com/tomo835g/Superconductors}{link}.

\begin{acknowledgments}
The author thanks Hodaka Kurokawa for his suggestion to tackle band gap estimation and for finding the data, Ya Zhuo and Jakoah Brgoch for communication, the conference Deep Learning and Physics 2019 at Yukawa Institute of Theoretical Physics, and Yoshiaki Shimada, who is not an author, for very fruitful discussions through which the idea of rolling up the periodic table came up at National Institute of Information and Communications Technology Open Summit 2019.

\end{acknowledgments}
\bibliography{rpt}

\end{document}


\title{Supplementary Information: Deep-Learning Estimation of Band Gap with the Reading-Periodic-Table Method and Periodic Convolution Layer} 
\author{Tomohiko Konno}
\affiliation{National Institute of Information and Communications Technology}
\email{tomo.konno@nict.go.jp}
\maketitle

\section{Supplementary Information}
\subsection{Averaged Scores}
As the previous work reported on single-shot scores, this study also reports on single-shot scores in the main text for a fair comparison. As the splitting of the dataset into training and test data is random, the scores vary depending on the split. Herein, we would like to report the average scores to evaluate the model.
The scores in the binary classification regarding whether the materials are nonmetals and the regression of the band gap are summarized in Tables~\ref{tab: binary-av} and ~\ref{tab: reg-av}, respectively. They represent the average scores over 56 and 80 models, respectively. The scores are comparatively on par with those obtained in the previous study, although our current scores are average values. 

\begin{table}[htbp]
\centering
\caption{Binary classification of band gap existence}\label{tab: binary-av}
\begin{tabular}{|c|c|c|c|c|}
\hline
accuracy & precision & recall & f1  & AUC \\ \hline
0.88 & 0.88 & 0.89 & 0.89 & 0.95 \\ \hline
\end{tabular}
\end{table}

\begin{table}[htbp]
\centering
\caption{Regression of band gap}\label{tab: reg-av}
\begin{tabular}{|c|c|}
\hline
R-squared & RMSE\\ \hline
0.87 & 0.55 \\ \hline
\end{tabular}
\end{table}

\subsection{Periodic Table with and without the Periodicity }
Herein, we provide a comparison between the original method and the extended method of reading periodic table with periodicity behind the table to check which method performs better. The parameters are identical for both methods, with the exception of the periodicity. However, we cannot disregard the fact that the number of weights in the network differs. Moreover, the scores are averaged. In the model without periodicity, for the classification, the accuracy was 0.88, precision was 0.88, recall was 0.90, f1 score was 0.89, and AUC was 0.95, and for the regression, the R-squared value was 0.88 and the RMSE was 0.53. These were the average values for 56 and 80 models, respectively. The results do not differ significantly; however, if we consider a problem relevant to the elements on both the right and left columns in the periodic table, and we use the temporally separate train--test data, we may obtain better results when using the method of reading periodic table with periodicity.

\bibliography{rpt}
